\documentclass[letterpaper]{article} 
\usepackage{aaai2027}  
\usepackage[hyphens]{url}  
\usepackage{graphicx} 
\usepackage{natbib}  
\usepackage{caption} 
\usepackage{algorithm}
\usepackage{algorithmic}

\usepackage{newfloat}
\usepackage{listings}
\DeclareCaptionStyle{ruled}{labelfont=normalfont,labelsep=colon,strut=off} 
\floatstyle{ruled}
\newfloat{listing}{tb}{lst}{}
\floatname{listing}{Listing}

\usepackage{booktabs}

\usepackage{tikz}
\usetikzlibrary{arrows.meta,positioning}
\newcommand{\op}{\diamond}

\usepackage{amsmath}
\usepackage{amssymb}

\nocopyright

\title{ALPS: Measuring Valid Creativity in Large Language Models with Mathematical Construction}
\author{
    Eric Xie, Wenqian Ye, Aidong Zhang
}
\affiliations{
    University of Virginia\\

    \{jrg4wx, pvc7hs, aidong\}@virginia.edu
}

\begin{document}

\maketitle

\begin{abstract}
Large language models produce outputs presented as discoveries — new proofs, conjectures, or molecules. Whether such an output that appears creative is truly original and effective is hard to establish: open-ended outputs require subjective judgment, the output may replicate something seen in training, or the task may be too simple to need creativity. We present ALPS (Austin-Law Proof-Synthesis), a benchmark that designs a task to measure \emph{valid creativity}: producing a solution that is original and can be proven correct. Each instance is a single equational law, certified to require either the construction of an infinite mathematical structure satisfying the law, or a proof that no such structure exists. Submissions are verified by automated proof checking with no human involvement, and a public generator produces new instances without limit, so LLMs are never evaluated on problems they may have seen. A portfolio of eight configurations of leading automated provers resolves 2.2\% of the 4,141-law evaluation pool, and a twentyfold budget increase adds 0.6\%: the obstacle is not compute, but the absence of any method that produces the tailored structure each law requires. Under a fixed protocol, the strongest reasoning model we test succeeds in 14\% of instances on the proof side, but none on the construction side. The remaining 97.2\% of the pool is unresolved at every configuration and budget we test. We release ALPS in full: the corpus, the generator, and the automated judge.

\end{abstract}

\begin{links}
    \link{Code}{https://github.com/EricX22/ALPS}
\end{links}

\section{Introduction}

As large language models (LLMs) begin to take on scientific work in the real world --- formalizing and proving theorems \citep{yang2023leandojo, xin2024deepseek}, proposing conjectures \citep{romera2024mathematical, gottweis2026accelerating, lu2024ai}, and designing molecules and materials \citep{jumper2021highly, merchant2023scaling} --- AI-for-science has grown increasingly interested in measuring an LLM's ability to reason and produce creative solutions \citep{chen2025ai4research}. Results within mathematics have shown this interest to be well-founded. In 2026, a general-purpose reasoning model resolved the roughly eighty-year-old Erdős unit-distance conjecture, a long-open question in discrete geometry \citep{openai2026unitdistance}. Producing such a result requires creativity, defined in the literature by two joint criteria: a creative solution must be original, and it must be effective, meaning it provides value for the task it addresses \citep{runco2012standard}. A solution that appears creative is not automatically a valid one: apparent novelty may be a repetition of something seen in training, and apparent effectiveness is unestablished until it is demonstrated. Two conditions must therefore hold: creativity was necessary to solve the problem, and the answer can be soundly justified. We call this combination \emph{valid creativity}. 


\begin{table}[t]
\centering
\small
\setlength{\tabcolsep}{5pt}
\begin{tabular}{@{}lccc@{}}
\toprule
 & Verifiable & Renewable & Constructive \\
 & \scriptsize\textit{no subjectivity} & \scriptsize\textit{no contamination} & \scriptsize\textit{no set procedure} \\
\midrule
Static problem sets      & \checkmark &            & \checkmark \\
\quad{\scriptsize MATH, code, miniF2F} & & & \\[1pt]
Synthetic puzzles        & \checkmark & \checkmark &            \\
\quad{\scriptsize random SAT, ETP}     & & & \\[1pt]
Open discovery           &            & \checkmark & \checkmark \\
\quad{\scriptsize hypotheses, molecules} & & & \\[1pt]
\midrule
\textbf{ALPS (ours)}     & \checkmark & \checkmark & \checkmark \\
\bottomrule
\end{tabular}
\caption{Each property removes one confound in measuring valid creativity: verification removes subjective judging, renewal removes contamination, and construction removes solution by any instance-independent method. Each prior benchmark family \citep{hendrycks2021measuring, chen2021evaluating, zheng2021minif2f, selman1996generating, bolan2025equational, gottweis2026accelerating, jumper2021highly} fulfills two of the three, but measuring valid creativity requires all three.}
\label{tab:properties}
\end{table}

Current evaluation methods fail to reliably measure the validity of a creative output. For an evaluation to be effective, the solution must be \textit{verifiable}, with an answer that can be checked objectively rather than judged. The problem supply must be \textit{renewable}; new instances can be generated automatically and without limit, so evaluation can always use problems created after any training cutoff. And the task must be \textit{constructive}, requiring solutions tailored to each instance. A benchmark admits a confound for each of the properties it lacks. Without renewability and constructiveness, creativity may not have been necessary: the problem could have been answered from memory or by a known general procedure. As depicted in Table~\ref{tab:properties}, every prior benchmark family has at most two of the three. Synthetic puzzles such as random SAT \citep{selman1996generating} and the Equational Theories Project \citep{bolan2025equational} are verifiable and renewable but can be solved with a general procedure given enough compute, so success measures the ability to execute a known method. Static corpora such as MATH \citep{hendrycks2021measuring}, coding benchmarks \citep{chen2021evaluating}, and miniF2F \citep{zheng2021minif2f} ask for constructed, checkable solutions, but draw from fixed sets that eventually contaminate training data, so a high score may reflect memorization \citep{xu2024benchmark}. Open scientific discovery, such as hypothesis generation \citep{gottweis2026accelerating, lu2024ai} and molecule design \citep{jumper2021highly}, demands construction and never exhausts its supply, but its answers are judged subjectively, so scores depend on the judge.

Formal mathematics is the one setting where each of these three properties can be fulfilled simultaneously. A formal proof can be checked automatically, and problems can be generated without limit, so verifiability and renewability come immediately. Construction is the property that must be established by design. Equational theory \citep{birkhoff1935structure} studies \textit{magmas}: sets with a single binary operation $\op$, and their \textit{laws}, universally quantified equations such as commutativity $x \op y = y \op x$ \citep{burris1981course, baader1998term}. A magma satisfying a law is a \textit{model} of it. The one-element magma, called the \textit{trivial} magma, models every law and carries no information. The substantive question is whether a law has a \textit{nontrivial} model, one with at least two elements. Within a finite carrier, such a model can always be found by search. Some laws, however, have no nontrivial finite model, yet have a nontrivial model all the same, necessarily infinite: these are the \textit{Austin} laws \citep{kisielewicz1997austin, bolan2025equational}. No finite search reaches an infinite model; it must be constructed and then justified.

We introduce the Austin-Law Proof-Synthesis (ALPS) benchmark. Each instance is a law with no nontrivial finite model, a property established in advance, which leaves exactly two possibilities: the law is trivial, collapsing every model so that $x=y$, or it is Austin, with an infinite model that satisfies it. Solving an instance means deciding which holds and proving it: a deductive proof on the trivial side, a constructed infinite model on the Austin side. Every answer is settled by an automatically validated proof certificate, so grading is objective and needs no human judge. The instances come from a generator that proposes candidate laws without limit, then screens each to keep only those with this property, so ALPS never runs short of fresh, unseen problems.

Austin laws sit where AI-for-science and mathematical provers are converging. The Erdős unit-distance conjecture was settled by a reasoning model through construction within an infinite space, at the frontier of what current language models can do \citep{openai2026unitdistance}. Computational mathematical provers approach from the other side with new methods of constructing infinite models \citep{janota2026case}. Austin laws are a renewable supply of this challenge: each is an infinite-construction problem with a definite answer, and fresh ones can be generated without limit.

\begin{table}[t]
\centering
\small
\begin{tabular}{@{}lp{3.1cm}p{3.1cm}@{}}
\toprule
 & Molecule design & ALPS instance \\
\midrule
Specify   & target profile: bind the pocket, non-toxic, synthesizable
          & a law $L$ the operation must satisfy on every assignment \\
\midrule
Construct & propose a structure satisfying all constraints
          & propose a magma satisfying $L$ \\
\midrule
Verify    & synthesis and assay confirm the profile
          & a proof certificate confirms the model \\
\bottomrule
\end{tabular}
\caption{Discovery proceeds by the same three phases in either domain;
in ALPS, every phase is machine-checkable.}
\label{tab:introduction}
\end{table}

What ALPS requires of the solver reaches well beyond equational theory. Resolving a law means proposing a new mathematical structure and establishing its global properties: the law $L$ specifies the conditions the operation must satisfy on every assignment of elements, the solver constructs a magma meeting them, and a proof certificate verifies that it does, across the infinitely many cases the solver can never check directly. Constructing a mathematical object that provably meets such a specification is itself an act of scientific discovery, and these three phases --- specify, construct, and verify --- recur across the sciences (Table~\ref{tab:introduction}). For example, in molecule design, they are a target profile, a proposed structure, and a synthesis assay; in ALPS, the law $L$, a magma, and a proof certificate. Because that proof decides both conditions of valid creativity from the answer alone, a verified solution is direct evidence that the language model created a novel structure and justified it. The claim is confined to what is certifiable: the problem admitted no solution by memory or routine, and the answer is provably correct --- properties of the problem and of the answer rather than of the process that produced them.

We demonstrate empirically that the task ALPS defines is constructive. A portfolio built from eight configurations of the strongest automated provers \citep{kovacs2013first, schulz2019faster, smallbone2021twee} resolves 2.2\% of the 4,141 laws that survive screening, and a twentyfold increase in per-prover budget adds 0.6\%. The remaining laws yield neither a model nor a triviality proof under any configuration we test, although each is certified to have a determinate answer; they are limited by the methods available rather than by the compute allotted to them. In full, our contributions are:
\begin{itemize}
\item a two-sided construction task on Austin laws, with every instance certified to have a determinate answer;
\item a corpus of such laws and a public generator that mints fresh instances without limit;
\item an automated judge that certifies submissions through independent channels --- a Lean proof \citep{moura2021lean} for triviality, and an automated-prover certificate \citep{kovacs2013first} for constructed models --- with no human in the loop; and
\item a strong automated baseline that establishes the method-bound hard tier and, along the way, settles an order-5 case left open by the Equational Theories Project (ETP) \citep{bolan2025equational}.
\end{itemize}

\section{Related Works}

Existing benchmark families each contain two of the three properties and fall short on the third (Table~\ref{tab:properties}). Static problem sets span competition mathematics \citep{cobbe2021training, hendrycks2021measuring}, graduate science \citep{rein2023gpqa}, coding \citep{chen2021evaluating}, and formal proof \citep{zheng2021minif2f, azerbayev2023proofnet, tsoukalas2024putnambench}. These are scored objectively, but fixed evaluation sets eventually enter training data, after which performance may reflect memorization \citep{xu2024benchmark, chen2025recent}. Holding problems secret \citep{glazer2024frontiermath} or releasing them over time \citep{white2025livebench, jain2025livecodebench} delays contamination but remains limited by the rate at which humans author new problems. Synthetic puzzles such as random SAT \citep{selman1996generating}, graph-generated reasoning tasks \citep{zhu2024dyval}, Sudoku variants \citep{seely2025sudoku}, and the equational laws of the Equational Theories Project \citep{bolan2025equational} renew without limit, but the regularity that makes their problems generable also makes them solvable by a known procedure, so success can measure execution of a method rather than construction of a solution. Open scientific discovery tasks such as hypothesis generation and molecule or materials design \citep{lu2024ai, gottweis2026accelerating, alkan2025survey, romera2024mathematical, merchant2023scaling, jumper2021highly} demand construction and never exhaust their supply, but their results are validated by human expertise or physical experiment, so evaluation is subjective and slow. ALPS combines the strengths of the three families: objective scoring, unlimited renewal, and a constructive task that no known general procedure solves.

ALPS grades answers by machine verification, made possible by theorem provers. Lean \citep{moura2021lean} is a proof assistant whose kernel mechanically checks each submitted proof, so a proof it accepts is correct by construction. A statement quantified over infinitely many inputs, such as a law holding under every assignment, is checked once, symbolically, rather than case by case. Finding proofs, unlike checking them, is difficult. Classical automated theorem provers such as Vampire and E \citep{kovacs2013first, schulz2019faster} search for first-order proofs by superposition \citep{bachmair1994rewrite}, repeatedly deriving consequences of the input until a contradiction is reached or no new consequences remain. More recently, language models have been trained to construct proofs in systems such as Lean \citep{yang2023leandojo, xin2024deepseek, hubert2026olympiad}, measured on formal proof benchmarks \citep{zheng2021minif2f}.

Equational theory studies the identities that operations satisfy and the entailments among them. The Equational Theories Project \citep{bolan2025equational} enumerates the implications among thousands of small laws and settles them automatically with provers and finite-model finders, singling out the Austin laws \citep{kisielewicz1997austin} --- those whose only nontrivial models are infinite --- as the hardest to resolve, since neither a finite-model search nor a short derivation settles them. Among methods tailored to equational laws, Paradox and Mace4 \citep{claessen2003new, mccune2003mace4} find finite models when they exist, Infinox \citep{claessen2011automated} proves that no finite model exists, and Twee \citep{smallbone2021twee} implements unfailing Knuth--Bendix completion \citep{knuth1970simple, bachmair1989completion}, orienting equations into rewrite systems. When a saturation halts without reaching a contradiction, its accumulated clauses describe a model; \citet{janota2026case} recently extracted an explicit model from such a saturation in a case study.

\section{The Austin-Law Proof-Synthesis Benchmark}

ALPS is built in three parts: a two-sided task, a corpus generator, and an automated judge. Each part provides one property listed in Table~\ref{tab:properties}: the admissibility certificate makes every instance a construction problem with a determinate answer, the extension engine renews the supply without limit, and the judge verifies answers automatically.

\subsection{Problem Formulation}

A magma is a set $M$, called the carrier, together with a single binary operation ${\op}\colon M \times M \to M$ and no further axioms. An equational law is a universally quantified identity between two terms built from variables and $\op$. Its order is the number of times the operation appears, so the commutative law $x \op y = y \op x$ has order two. A magma \textit{satisfies} a law when the identity holds under every assignment of carrier elements to its variables; such a magma is called a \textit{model} of the law.

Every ALPS instance is a law of the form $L\colon x = T[x, y, z, \dots]$: a designated variable $x$, isolated on the left-hand side, equated with a term $T$ of order five or more, where brackets list variables from which $T$ is built; $x$ may itself recur within $T$. The trivial magma contains only one element, and is thus a model of every law. A model is nontrivial when its carrier holds at least two distinct elements. The question ALPS poses for each $L$ is whether a nontrivial model exists at all: either $L$ entails $x = y$, collapsing every model to a single value, or some nontrivial magma satisfies it.

A nontrivial model, when one exists, may be finite or infinite. The finite case is a matter of search: enumerating operations on carriers of increasing size must eventually find it, requiring none of the law-specific creativity the benchmark aims to measure. We therefore admit a law into the benchmark only when it carries a machine-checked proof that no nontrivial finite model exists, calling such laws \textit{admissible}. Admissibility forces a dichotomy: every law either entails $x = y$ or it does not, so each instance has a determinate answer. Since laws are universally quantified, $L \models x = y$ states that any two elements of any model are equal. For an admissible law, the two cases are exactly
\begin{equation}
    \label{eq:dichotomy}
    \underbrace{L \models x = y}_{\text{trivial}}
    \quad\text{or}\quad
    \underbrace{L \text{ has an infinite nontrivial model}}_{\text{Austin}}.
\end{equation}
Laws of the second kind are Austin laws \citep{kisielewicz1997austin}.

The asymmetry between the two sides of the task is essential to what the benchmark measures. Whether $L$ entails $x = y$ is a question of first-order logic, the language of statements built from the operation $\op$, equality, logical connectives, and quantifiers ranging over carrier elements \citep{baader1998term}. Proof search within first-order logic is complete: if the entailment holds, a systematic search will eventually confirm it. The trivial side is, in principle, settled by deduction. The Austin case admits no such guarantee. A search that fails to derive $x = y$, however long it runs, yields no model. Because equational entailment is undecidable in general \citep{baader1998term}, no uniform procedure can be expected to close the gap. Establishing that a law is Austin means inventing an infinite magma tailored to the individual law to satisfy it. This synthesis, which deductive search does not supply, is the capability that ALPS isolates. 

\begin{table}[t]
\centering
\small
\begin{tabular}{@{}lrr@{}}
\toprule
Screening stage & Removed & Remaining \\
\midrule
Generated candidates  & ---      & 10{,}474 \\
Finite-model filter   & 1{,}085  & 9{,}389  \\
Admissibility prover  & 1{,}906  & 7{,}483  \\
\midrule
\multicolumn{3}{@{}l}{Admissible pool:} \\
\quad Proven trivial            & & 3{,}080 \\
\quad Austin classes (262 laws) & & 195     \\
\quad Unresolved residual       & & 4{,}141 \\
\bottomrule
\end{tabular}
\caption{The screening pipeline. In the top block, each stage removes the number of candidates shown and passes the remainder to the next. The bottom block partitions the certified pool by the proofs screening produced: 3,080 proven trivial, 262 proven Austin (reported as 195 equivalence classes), and 4,141 unresolved by screening, summing to 7,483.}
\label{tab:funnel}
\end{table}

\subsection{Corpus Construction}
\textbf{Generation by extension.} Admissible laws are too rare for random sampling to produce in quantity. We instead grow the corpus outward from laws already proven to be Austin: a seed law $x = T$ is extended by selecting any variable $v$ in $T$ and replacing one occurrence with $v \op w$ or $w \op v$ for some other variable $w$. Because a candidate inherits most of the seed's term structure, it often inherits the mechanism that prevents a finite model, increasing the likelihood of finding new admissible laws relative to random sampling. The construction is recursive --- laws confirmed Austin at order $n$ seed the round at order $n+1$ --- so the corpus has no terminal order. This keeps the supply of new laws renewable: fresh instances can be minted after any training cutoff, so evaluation never depends on problems a model may have seen.  

\textbf{Admissibility.} Each candidate then enters the screening pipeline (Table~\ref{tab:funnel}). A finite-model filter discards every candidate it can satisfy nontrivially. The survivors face the admissibility prover, which attempts a proof that no nontrivial finite model exists. The claim cannot be posed to the prover directly, because finiteness is not expressible in first-order logic; the proof instead works through a consequence of finiteness: on a finite carrier, an injective map is also surjective. Let $S$ be a subterm of $T$ containing every occurrence of $x$, so that the law reads $x = C[\,S(x,\bar y)\,]$, where the surrounding context $C$ contains no $x$, and $\bar y$ abbreviates the remaining variables. The law itself thus makes $C$ a left inverse of $S$, so $S$ is injective in every model. Every finite model of $L$ additionally satisfies the first-order statement

\begin{equation}\label{eq:surj}
\mathrm{surj}(S)\colon\quad
\forall \bar y\,\forall u\,\exists x\colon\; S(x,\bar y) = u.
\end{equation}

The admissibility query submits both facts to the prover as a single derivation task, written $\vdash$ in contrast to the entailment $\models$ of Equation~\eqref{eq:dichotomy}:

\begin{equation}\label{eq:admquery}
\{L,\ \mathrm{surj}(S)\} \vdash x = y.
\end{equation}

A refutation proof of Equation~\eqref{eq:admquery} is the admissibility certificate: a derived statement holds in every model of its premises, and every finite model of $L$ satisfies both premises, so every finite model of $L$ is trivial. As a result, either $L$ entails $x = y$, or $L$ has a nontrivial model, which must then be infinite. Admission requires this certificate, and none exists for a law with a nontrivial finite model. When injectivity cannot be proven, the candidate remains uncertified and is discarded even though it may have been admissible. The screening provers classify the admissible pool into laws proven trivial, laws proven Austin, and the residual that the baseline of the next section attempts.

\textbf{Deduplication.} Because every law descends from a seed set, an extension may pose the same problem as its seed or a sibling, defined as when each entails the other:
\begin{equation}\label{eq:equiv}
L_1 \models L_2
\quad\text{and}\quad
L_2 \models L_1,
\end{equation}
in which case the two laws have the same models and are called equivalent. A pair is distinct when either direction of Equation~\eqref{eq:equiv} fails, and a failure is witnessed by a separating model: a magma that is a model of $L_1$ but not of $L_2$ shows $L_1 \not\models L_2$. The models constructed during screening double as these witnesses. Across the 34,191 pairs among the 262 laws proved Austin, 33,936 pairs are separated by a model found during screening. The prover settles the remainder by deriving both directions of Equation~\eqref{eq:equiv}: 255 pairs are proved equivalent, and no pair is left undecided. Because equivalence is transitive, the 255 proved equivalences chain into groups of mutually equivalent laws, each counted as a single problem. A group of $k$ laws accounts for $k(k-1)/2$ of the proved pairs but only $k-1$ duplicates, which is why 255 pairs merge away only 67 laws: the 262 laws collapse to 195 classes. We report corpus size in classes, so the count reflects only distinct problems and no solution is credited twice.

\begin{figure*}[t]
\centering
\begin{tikzpicture}[
  font=\footnotesize,
  box/.style={draw=black!50, fill=black!3, rounded corners=2pt, align=flush left,
              text width=7.5cm, inner sep=6pt},
  boxb/.style={draw=blue!50!black!60, fill=blue!5, rounded corners=2pt, align=flush left,
               text width=7.5cm, inner sep=6pt},
  boxg/.style={draw=green!40!black!60, fill=green!5, rounded corners=2pt, align=flush left,
               text width=7.5cm, inner sep=6pt},
  boxr/.style={draw=red!45!black!60, fill=red!4, rounded corners=2pt, align=flush left,
               text width=7.5cm, inner sep=6pt}
]
\node[box, anchor=north west] (l1) at (0,0) {%
\textbf{A toy inadmissible law}\\[1pt]
$L_0:\;\; x = x \op (y \op y)$};
\node[boxb, anchor=north west] (l2) at ([yshift=-2.5mm]l1.south west) {%
\textbf{Sample submission}\\[1pt]
$E_0:\; u \op v = u$\\
{\scriptsize\itshape the operation returns its left argument}};
\node[boxg, anchor=north west] (l3) at ([yshift=-2.5mm]l2.south west) {%
\textbf{The two checks}\\[2pt]
(a) $E_0 \vdash L_0$: rewriting with $u \op v = u$ ($u = x$, $v = y \op y$):\\
\phantom{(a) }$x \op (y \op y) = x$, so both sides of $L_0$ agree. \hfill $\checkmark$\\[2pt]
(b) $E_0 \cup \{a \neq b\}$ satisfiable: the two-element magma with\\
\phantom{(b) }$s \op t = s$ for all $s,t$ satisfies $E_0$ and keeps $a \neq b$. \hfill $\checkmark$\\[2pt]
Certified: $E_0$ has a nontrivial model satisfying $L_0$.};
\node[box, anchor=north west] (r1) at (8.1,0) {%
\textbf{A hard-tier law}\\[1pt]
$L:\;\; x = y \op (x \op (((w \op y) \op ((y \op z) \op (y \op x))) \op y))$};
\node[boxb, anchor=north west] (r2) at ([yshift=-2.5mm]r1.south west) {%
\textbf{o4-mini's submission}\\[1pt]
$E_1:\; y \op (x \op y) = x \qquad E_2:\; x \op (z \op y) = x \op y$\\
{\scriptsize\itshape $z$ is universally quantified and may be instantiated at any term}};
\node[boxr, anchor=north west] (r3) at ([yshift=-2.5mm]r2.south west) {%
\textbf{The two checks}\\[2pt]
(a) $E \vdash L$: instantiating $z$ at the law's middle subterm,\\
\phantom{(a) }$y \op (x \op (z \op y))
  \overset{E_2}{=} y \op (x \op y)
  \overset{E_1}{=} x$ \hfill $\checkmark$\\[2pt]
(b) $E \cup \{a \neq b\}$ satisfiable: for any $t$,
  $t \overset{E_1}{=} y \op (t \op y) \overset{E_2}{=} y \op y$,\\
\phantom{(b) }so every element equals $y \op y$: $E$ forces $a = b$. \hfill $\times$\\[2pt]
Rejected at check (b): $E$ admits no model with two distinct elements.};
\end{tikzpicture}
\caption{The two checks on two laws. Left: a sample submission on a toy, inadmissible law passes both checks; for admissible laws, admissibility forces the witnessing model to be infinite. Right: o4-mini's submission on a hard-tier law passes check (a) but fails check (b), the ``too-strong'' failure mode most common across the LLMs we evaluate.}
\label{fig:certs}
\end{figure*}

\subsection{Answer Verification}
Each solved ALPS instance is a machine-verified mathematical statement. The two sides of the task require different verification processes, so the judge issues a certificate through the appropriate channel: a proof checked by the Lean kernel for the trivial side \citep{moura2021lean}, and an automated-prover certificate for the constructed model \citep{kovacs2013first}. Both channels are fully automatic, and a submission is accepted only if its check succeeds. Verification rests on two trusted checkers, the Lean kernel on the trivial side and prover saturation on the construction side, cross-checkable by a second prover \citep{smallbone2021twee}.

\textbf{The trivial channel.} For each law, the judge generates the proposition
\begin{equation}\label{eq:trivialgoal}
\begin{split}
\textsc{TrivialGoal} \equiv\ &\forall M\ \forall {\op}\colon M \times M \to M,\\
&\mathrm{Law}(\op) \rightarrow \forall a, b : M,\ a = b,
\end{split}
\end{equation}
where $\mathrm{Law}(\op)$ asserts that $\op$ satisfies $L$. In words, Equation~\eqref{eq:trivialgoal} states that any magma satisfying $L$ has all of its elements equal: it is the trivial side of Equation~\eqref{eq:dichotomy}, with the magma quantified inside the statement so that a single proof covers every carrier and every operation at once. A submission is a Lean proof of Equation~\eqref{eq:trivialgoal}. The statement is generated by the judge, never written by the submitter: the judged file fixes Equation~\eqref{eq:trivialgoal} in a header, inserts the submitted proof, and requires it to elaborate at exactly that type, and the proof's axiom footprint must fall within a fixed allowlist, which blocks declared axioms and placeholder proofs that would allow a solver to postulate what it must prove. Under these constraints, a proof the kernel accepts is correct by construction \citep{moura2021lean}. In practice, such a proof formalizes a chain of equalities $a = t_1 = \cdots = t_k = b$ in which each step applies one instance of $L$.

\textbf{The construction channel.} A model of an Austin law has an infinite carrier, but a submission must be a finite object a machine can check. The solver instead submits a finite presentation $E$: a finite set of equations over $\op$, whose models are the magmas satisfying every equation in $E$. $E$ must satisfy two claims: (i) every model of $E$ satisfies $L$, and (ii) some model of $E$ has two distinct elements. The claims constrain $E$ from opposite sides: it must be strong enough to entail the law, yet weak enough that its models do not all collapse to one element. The judge accepts only when a prover certifies both. Figure~\ref{fig:certs} (left) demonstrates this process on a toy law. Formally, the two queries are

\begin{equation}\label{eq:construction}
\text{(a)}\;\; E \vdash L
\qquad\qquad
\text{(b)}\;\; E \cup \{\,a \neq b\,\} \text{ is satisfiable.}
\end{equation}

Check (a) is a derivation task of the same kind as Equation~\eqref{eq:admquery}: the prover returns a refutation proof that $L$ is derivable from the equations of $E$, and by the same reasoning as before, every model of $E$ satisfies $L$, which is claim (i). Check (b) asks whether some magma satisfies every equation of $E$ while keeping two elements $a$ and $b$ distinct. The prover answers by saturation: starting from $E \cup \{a \neq b\}$, it derives consequences until either a contradiction appears or nothing new can be derived, at which point the set is called saturated. A saturated set that contains no contradiction is satisfiable, provided no inference was skipped along the way, and its accumulated consequences describe a model \citep{bachmair1994rewrite}. Provers do skip inferences in some configurations, trading completeness for speed, and report when they have done so. The judge therefore accepts check (b) only when the prover reports that its strategy retained completeness. A terminating, complete saturation of $E \cup \{a \neq b\}$ thus establishes claim (ii). The two checks compose: by (ii) some magma satisfies $E$ with $a \neq b$, by (i) that magma satisfies $L$, so $L$ has a nontrivial model, and the admissibility certificate of the previous section forces that model to be infinite. $L$ is therefore Austin. Neither query mentions infinity: both are finite first-order tasks, and infinitude enters only through the admissibility certificate the instance already carries.

Although grading is fully automatic, the checks themselves reject any submission that does not encode a genuine construction, so a certificate cannot be earned without solving the instance. Submitting the law itself as $E$ gains nothing: check (a) becomes immediate, but check (b) reduces to the saturation of $L \cup \{a \neq b\}$, the computation that fails to terminate on these laws. A presentation too strong collapses every element and fails check (b) (Figure~\ref{fig:certs}, right); one too weak, such as the left projection $\{x \op y = x\}$, fails check (a). The judge is validated against controls of all three kinds before any evaluation run, and the submission format admits only unconditional equations, so implications, disequations, and literal restatements of $x = y$ never reach the prover.

A single Lean judge covering both sides is not currently possible. The model families that are straightforward to formalize in Lean are arithmetic. No model of an admissible law is among them, since an arithmetic operation satisfying the law would yield a nontrivial finite model, contradicting admissibility. The models that do exist are built from saturations \citep{janota2026case}, and formalizing these in Lean awaits a proof of ground confluence. The construction side is therefore graded by the prover certificate.

\section{Experiments}

\subsection{The Automated Frontier}

We establish what automation resolves using a portfolio spanning eight configurations of leading off-the-shelf provers: Vampire 5.0.1 in five modes (proof search and saturation, under Knuth--Bendix and lexicographic path orderings) \citep{kovacs2013first}, the E prover in a proof and a saturation mode \citep{schulz2019faster}, and Twee 2.6.1, an implementation of unfailing completion built for equational reasoning \citep{smallbone2021twee}, each running over per-prover budgets of up to 600 seconds. A law is resolved if any configuration returns a certificate. Finite-model finders such as Mace4 \citep{mccune2003mace4} and Paradox \citep{claessen2003new} cannot succeed and are omitted, as our admissibility classification certifies that no nontrivial finite model exists. Saturation and completion are automated methods in the portfolio that can exhibit a model: a saturation that terminates without contradiction describes one (Section 3.3).

Sweeping the portfolio over the admissible pool distinguishes two populations. The models of the 262 laws proved Austin during screening are constructed in a median of 0.1 seconds, and sweeping the 4,141-law residual with the full portfolio reclassifies a further 114 laws (2.8\% of the residual): 110 triviality proofs and 4 completion-only models. The remaining 4,027 laws resolve under no configuration at any budget up to 600 seconds and constitute the hard tier, defined by this gap in the resolution-time distribution. If this gap were simply a result of insufficient compute, additional budget would be expected to continue resolving new laws of both kinds at a comparable rate. Instead, of the 114 residual laws the sweep resolves, 91 resolve at the 30-second per-configuration budget. Raising the budget by a factor of 20 resolves only 23 more, each one a triviality proof (Table~\ref{tab:frontier}). All four models the sweep recovers, together with the 262 constructed during screening, are identified within the first 30 seconds.

The resolved tier contains a result of independent interest. Among the laws the portfolio resolves is an order-five law and its dual (12857 and 33436 in the ETP numbering) that the Equational Theories Project recorded as having only trivial finite models, leaving the existence of an infinite model open \citep{bolan2025equational}. Two independent methods, Vampire saturation and Twee completion, terminate without deriving $x = y$. As established in Section 3.3, a terminating, complete run describes a model where $x = y$ fails, and admissibility forces that model to be infinite. We verify the extracted presentation, a set of 27 equations, directly: it satisfies the law and does not force $x = y$. The model is of the saturation-derived kind described in Section 3.3, so its Lean formalization awaits the ground-confluence proof noted there; the full worked example is provided in the Appendix.

\begin{table}[t]
\centering
\begin{tabular}{@{}rrrr@{}}
\toprule
Budget (s) & New models & New trivial & Unresolved \\
\midrule
30  & 4 & 87 & 4,050 \\
60  & 0 & 8  & 4,042 \\
120 & 0 & 5  & 4,037 \\
300 & 0 & 4  & 4,033 \\
600 & 0 & 6  & 4,027 \\
\bottomrule
\end{tabular}
\caption{Newly resolved laws at each per-configuration budget over the 4,141-law residual. All models appear at 30 seconds; every later resolution is a triviality proof, and 4,027 laws (97.2\%) remain unresolved.}
\label{tab:frontier}
\end{table}

\subsection{The LLM Frontier}
 
Language models enter ALPS on the same terms as the automated portfolio: nothing in the task references how an answer is produced, only that it certifies. We evaluate them under a fixed protocol designed so that failures are attributable. Every solver receives identical inputs --- the law, the task statement, and a fixed answer format --- and may submit to the judge up to three times, receiving each verdict as feedback. On the trivial side, the formalization burden is removed. The LLM emits the entire chain of equalities between two arbitrary carrier elements $a$ and $b$, each step being one application of the law, and the autoformalization harness matches each step to assemble the Lean proof itself, so the task is focused on finding the derivation rather than writing Lean. This assistance is fixed, and the harness is frozen across LLMs. A \textit{waypoint lemma} is an equality a prover has already derived from the law. Provided waypoints, the LLM does not need to plan the whole chain at once: it can build a shorter chain to a waypoint and continue from there. Waypoints cannot fix an individual step, since every step the LLM writes must still be a correct application of the law. Comparing runs with and without waypoints therefore shows where failures lie: in finding chains, which waypoints make easier, or in writing correct steps, which waypoints do not affect. Saturation provers are withheld, since with them a solver could simply re-run the portfolio; the judge and the harness are the tools available to any solver built on ALPS.

We construct one law set for each of the two sides of the task. The certified-easy set tests whether an LLM can produce a valid solution under the most favorable conditions the corpus offers: 63 trivial-tier laws with simple collapse chains $a = t_1 = \cdots = t_k = b$. Rewriting from both ends until the two chains meet resolves all 63 in under five seconds, and most chains are only three law applications long. The construction experiment is posed over a sample of 25 of the 4,027 ``hard-tier'' laws that no automated configuration resolves. These laws are open in both directions so we evaluate both channels: the construction channel of Section 3.3, and the trivial channel without waypoints. Both evaluation sets contain only laws minted by the extension procedure, and the generator runs past the released corpus, so future evaluations can draw instances created after any LLM's training cutoff.

 \begin{table}[t]
\centering
\begin{tabular}{@{}lcccc@{}}
\toprule
 & \multicolumn{3}{c}{Certified-easy ($n=63$)} & Hard ($n=25$) \\
\cmidrule(lr){2-4}
LLM & Default & Waypoints & $\downarrow$ Effort & Construction \\
\midrule
GPT-4.1 & 0 & 0 & ---     & 0 \\
o4-mini & 0 & 0 & 0 & 0 \\
o3      & 6 & 9 & 0 & 0 \\
\bottomrule
\end{tabular}
\caption{Verified LLM solves under the fixed protocol. Waypoints are prover-derived lemmas included in the prompt. Reasoning effort is medium except where marked $\downarrow$, which denotes low effort; GPT-4.1 has no effort setting.}
\label{tab:llm}
\end{table}

Table~\ref{tab:llm} reports the trivial-side results. Without waypoints, o3 at medium reasoning effort produces six verified solutions on the 63 certified-easy laws (9.5\% pass@1), and nine (14\%) with waypoints. GPT-4.1 and o4-mini produce none even with waypoints, and setting o3's reasoning effort to low reduces its solves to zero. The verified chains are three to four law applications long, most are accepted on the second or third feedback round, and a solved law costs a median of roughly 50,000 completion tokens. Failures share one dominant pattern at every setting: the submitted chains contain steps that are not applications of the law --- no substitution instance of either side of $L$ rewrites one term into the next --- and the harness rejects the chain at the matching stage.

Solve events are stochastic. A second run using o3 under identical settings reproduces one of the nine solves. Single-run counts are therefore pass@1 samples rather than a fixed solved set. The stable signals are binary: o3 produces verified solves in every run, GPT-4.1 and o4-mini in none, and low reasoning effort in none. The variation is expected: each run samples a fresh reasoning trace, and the harness provides minimal guidance beyond the three feedback rounds, so these results are a lower bound on language-model performance. The waypoint comparison sits within this variation: given the low reproduction rate, the difference between six and nine is not distinguishable from run noise, and we treat the effect of waypoints as suggestive. Its smallness is consistent with the dominant failure mode: rejected chains fail on invalid steps, which waypoints do not repair.
 
o3, o4-mini, and GPT-4.1 each attempt the 25 hard-tier laws over three feedback rounds, and no submission passes both checks. On the same laws, the trivial channel without waypoints also produces no verified solution, so every sampled law remains open in both directions. Classifying each law by its final-round submission, 23 of o3's 25 presentations are overly strong: they entail the law but admit only one-element models, passing check (a) and failing check (b). The remaining two are too weak, maintaining two distinct elements but failing to entail the law. o4-mini divides into 11 that are too strong, 4 that are too weak, and 10 that fail both checks; GPT-4.1 divides into 10, 11, and 4. Figure~\ref{fig:certs} (right) shows one such submission. On final submissions, o3 clears exactly one check on all 25 laws, GPT-4.1 on 21, and o4-mini on 15. The failure profile distinguishes the LLMs despite all submissions failing: o3's failures concentrate in a single mode, while the weaker LLMs spread across all three. Producing a presentation that is neither too strong nor too weak is the capability the construction side tests.

\section{Discussion}
The solver failures of Section 4.2 reveal weaknesses that are not specific to mathematical reasoning. Trivial-side failures concentrate in step validity rather than planning, which self-verification at each step could address. Construction-side submissions fail as too strong or too weak, with the judge reporting which. Passing both checks requires balancing two opposing requirements: satisfying a global specification without collapsing into a degenerate solution. Both weaknesses concern creative generation under constraints, a requirement of general scientific problem-solving, not just equational theory. These patterns point to a verify-and-refine loop, one that drafts a structure to satisfy global constraints, reads a verdict, and revises accordingly. If these weaknesses generalize, other domains should have arrived at the same remedy independently; software synthesis did, standardizing it as counterexample-guided inductive synthesis (CEGIS) \citep{solar2006combinatorial}, the architecture underlying program sketching \citep{solar2013program}, syntax-guided synthesis \citep{alur2013syntax}, and invariant discovery \citep{garg2014ice}.

ALPS differs from those settings in one step. There, a human supplies the candidate space, as a sketch or a grammar; in ALPS, no candidate space is known (Section 3.3), so the proposal must come from the solver. That proposal step is the construct phase of the specify--construct--verify cycle (Table~\ref{tab:introduction}), so the capability ALPS motivates extends wherever a specification can be stated and a solution checked. Because each run leaves a certified failure profile and the corpus renews, evaluation, analysis, and redesign can continue without exhausting the benchmark. The Appendix gives the full protocol prompts, per-configuration sweep results, and the complete failure classification.

\section{Conclusion}

We presented ALPS (Austin-Law Proof-Synthesis), a benchmark that measures valid creativity in language models: solving problems for which creativity is necessary, with answers that are soundly justified. Each instance asks the solver to construct an infinite mathematical structure satisfying a freshly minted law, or to prove that no such structure exists, and each is certified in advance to ensure finite search cannot settle it. Creativity is necessary because renewability and constructiveness remove the confounds: instances minted past any training cutoff cannot be answered from memory, and a task automation rarely resolves --- eight configurations of provers resolve 2.2\% of the 4,141-law pool, and a twentyfold budget increase adds 23 triviality proofs and no new models --- cannot be answered by routine. Justification is secured by verifiability: acceptance is decided by the machine-checked certificate alone. The strongest LLM we test verifies 14\% of the certified-easy set and constructs nothing on the hard tier, where it satisfies one of the two checks on every law and weaker LLMs sometimes satisfy neither. That failure profile specifies the verifier-guided solvers the benchmark rewards. We therefore release ALPS --- corpus, generator, and automated judge --- with 97.2\% of the evaluation pool unresolved by every configuration we test.

\bibliography{aaai2027}


%

\appendix

\lstset{%
  basicstyle={\footnotesize\ttfamily},
  breaklines=true, showstringspaces=false, frame=single,
  xleftmargin=0pt, columns=fullflexible,
  literate={◇}{{$\diamond$}}1 {—}{{---}}1 {–}{{--}}1
           {•}{{$\bullet$}}1 {→}{{$\rightarrow$}}1 {✓}{{$\checkmark$}}1
}

\renewcommand{\thetable}{A\arabic{table}}
\renewcommand{\thefigure}{A\arabic{figure}}
\setcounter{table}{0}
\setcounter{figure}{0}

\newpage
\section*{Appendix}

\section{Worked Example: Laws 12857 and 33436}
\label{app:worked}

Among the laws the portfolio resolves is a pair the Equational Theories
Project \citep{bolan2025equational} recorded as having only trivial finite
models, with the existence of an infinite model left unresolved. In the ETP numbering they are laws 12857 and
33436:
\begin{align*}
L_{12857}\colon\quad & x = y \op ((x \op (y \op (z \op z))) \op y), \\
L_{33436}\colon\quad & x = ((y \op (((z \op z) \op y) \op x)) \op y).
\end{align*}
The two are duals: reversing the arguments of every occurrence of $\op$ in one
yields the other. A magma satisfies one exactly when its opposite magma
satisfies the other, so the pair is a single open case up to duality, and
resolving either resolves both.

\subsection{Two independent certificates}

Both laws were resolved twice, by methods that share no search strategy.

Vampire 5.0.1 \citep{kovacs2013first} was run in ordered saturation mode (\texttt{-sa otter}, default
Knuth--Bendix ordering) on the law together with the assertion that two
elements are distinct. It terminated with \texttt{SZS status Satisfiable} in
7.95 seconds and 126\,MB of memory, printing an active set of 357 unit
equations. A terminating, complete saturation containing no contradiction
describes a model, as set out in Section~3.3; here the model is one in which
$x = y$ fails, and the admissibility certificate the instance already carries
forces that model to be infinite.

Twee 2.6.1 \citep{smallbone2021twee} was run independently on both laws and returned \texttt{SZS status
CounterSatisfiable} for each, deriving 55 rules for 12857 and 54 for 33436.
Unfailing completion is complete by construction, so this is a second
certificate for the same conclusion by a different route.

\subsection{Verifying the presentation}

The saturated presentation is checked directly rather than taken on the
prover's word. Three properties are confirmed. The ordering check verifies that
the ordering used to evaluate the presentation matches the one recorded in the
certificate header, which is necessary because ground confluence is inherited
from the saturation only with respect to the ordering the prover saturated
under; every pre-ordered equation is comparable under that ordering. The law
then holds on every ground instance tested, 27 of 27, and rewrites fire in all
of them, so the model is not satisfying the law vacuously. Finally the two
Skolem constants of the distinctness axiom normalise to different terms, which
establishes that the carrier has at least two elements.

The presentation is too large to reproduce here; the certificate exists in the released code, and \texttt{ordered\_model.py} reproduces all three
checks from it.

\subsection{Why this model is not Lean-checkable today}

The presentation is not an off-the-shelf term rewriting system, which is why
this model is graded by the prover certificate rather than by the Lean kernel
\citep{moura2021lean}.
Of the 357 equations, a substantial minority carry a free variable on each
side, so they cannot be oriented into rewrite rules in the usual sense.

That is a constraint on printing the system, not on the construction. The model
rewrites a ground term whenever the ground instance strictly decreases, and a
variable unbound on the other side of an equation may be mapped to any ground
term that makes the instance decrease --- in practice the smallest constant.
Every equation is therefore usable, in whichever direction decreases the
instance. Termination is immediate because the ordering is well founded on
ground terms, and ground confluence is inherited from saturation under
unfailing completion.

What is missing is an off-the-shelf certifier. Tools such as CSI and TTT2 check
plain rewriting systems, and the object here is an ordered one. A Lean
formalization of this model therefore awaits a machine-checked ground
confluence proof. Models of the algebraic kind, by contrast, are Lean-checkable
today; no admissible law admits one, for the reason given in Section~3.3.

\section{LLM Evaluation Protocol and Prompts}
\label{app:protocol}

\subsection{The fixed protocol}

Every solver receives the same three inputs: the law, the task statement, and a
fixed answer format. A submission may be sent to the judge up to three times,
and each verdict is returned verbatim as feedback before the next attempt. No
solver receives a hint that is not derived mechanically from the law itself,
and the harness is frozen across all language models, so differences in outcome
are attributable to the model rather than to the scaffolding.

Saturation provers are withheld from the language-model setting. With access to
them a solver could re-run the automated portfolio of Section~4.1 rather than
construct anything; the judge and the autoformalization harness are the tools
available to any solver built on ALPS.

All LLM calls use provider-default sampling through a single API gateway. The
APIs expose no seed parameter, and the reasoning LLMs are nondeterministic at
any setting, so bitwise reproduction of a run is not available. Run-to-run
variation is therefore measured rather than controlled: the reproduction run
of Section~4.2 repeats the nine solved and five matched unsolved laws under
the identical frozen harness. The released result files
record every attempt; the submitted chain is withheld from solved rows, since
a verified chain is a solution to a certified-easy law, and every reported
count is reproducible from the retained fields.

\subsection{The trivial channel}

On the trivial side the formalization burden is removed. The model emits a
chain of equalities between two arbitrary carrier elements $a$ and $b$, each
step being one application of the law, and returns it as JSON. The harness
matches each step against the law, bridges gaps of at most three single
applications automatically, and assembles the Lean proof itself. The model
therefore searches for a derivation rather than writing Lean.

A step is accepted only when some substitution instance of one side of the law
rewrites one term into the next at exactly one position, with everything
outside that position unchanged. Two rejections are absolute: relating $a$ and
$b$ directly, and any jump the bridging search cannot close.

A \emph{waypoint lemma} is an equality a prover has already derived from the
law, supplied in the prompt. Waypoints shorten planning --- the model can build
a chain to a waypoint and continue from there --- but they cannot repair an
individual step, since every step the model writes must still be a correct
application of the law. Comparing runs with and without waypoints therefore
separates failures of chain-finding from failures of step validity.

\subsection{The construction channel}

On the construction side the model proposes a finite presentation $E$: a set of
unconditional equations over $\op$, returned as JSON. The submission format
admits only unconditional equations, so implications, disequations and literal
restatements of $x = y$ never reach the prover. The judge then runs the two
checks of Equation~(6).

\subsection{Prompts}

Figure~\ref{fig:prompt-trivial} gives the trivial-channel prompt and
Figure~\ref{fig:prompt-construct} the construction-channel prompt, both
verbatim. Braces marked \texttt{\{law\}} and \texttt{\{hint\_block\}} are
substituted at call time; the hint block is empty in the no-waypoint condition
and lists the prover-derived waypoint lemmas otherwise. On a rejected
submission the harness appends the judge's reason to the next round's prompt:
\texttt{Your previous chain was REJECTED: \{feedback\}} on the trivial side,
and \texttt{Your previous proposal failed to certify: \{feedback\}} followed by
\texttt{Try a different presentation.} on the construction side.

\begin{figure*}[t]
\begin{lstlisting}
You are proving a magma law forces triviality. You will NOT write Lean — you
give only a sequence of terms, and a theorem prover checks each step.

Law:  {law}
Read it as a rewrite rule. Write the law as  L = R, where L is the single
variable on the left and R is the big right-hand term. ONE step does exactly
one of:
  • pick any subterm `e` of the current term and replace that ONE occurrence
    with R, taking L := e (so `e` becomes R with the law's left variable set
    to `e`); OR
  • the reverse — replace a subterm that exactly matches an instance of R by
    the corresponding `e`.
Everything OUTSIDE that one chosen subterm stays byte-for-byte identical.

Goal: build  a = t1 = t2 = ... = b. Adjacent terms should be only a FEW (1–3)
such steps apart — you may skip intermediate terms and the checker will fill
short gaps automatically, so give WAYPOINTS, not necessarily every atomic
step. Prefer smaller jumps.

HARD RULES — a jump breaking any of these is rejected, so self-check each one
before writing it:
  • You may NOT write "a = b" or otherwise relate a and b directly. They are
    opaque atoms; the only way to connect them is through the law.
  • Each jump must be a GENUINE short derivation — at most ~3 single law
    applications apart. A leap between unrelated terms cannot be bridged and
    is rejected.
  • Every application is a REAL instance: a subterm `e → R[L:=e]` (or its
    reverse) for some choice of the law's other variables. If you can't name
    that instance, the step is invalid.

STRATEGY: rewrite FORWARD from `a` (each rewrite grows the term via the law)
toward a term the law forces to collapse — these laws let you derive
`op(_, _) = (any element)`; once both `a` and `b` reduce to a common term, the
chain closes.{hint_block}

Return ONLY JSON:  {"chain": ["a", "<term>", ..., "b"]}   first "a", last "b";
use only a, b, ◇, ().
Example (toy law  x = y ◇ y):  {"chain": ["a", "(a ◇ a)", "b"]}
  step 1: `a` → `(a ◇ a)` — the law with x:=a, y:=a gives  a = a◇a. ✓
  step 2: `(a ◇ a)` → `b` — reverse: the law with x:=b, y:=a gives  b = a◇a,
    so  a◇a = b. ✓
Only write steps you can justify this way.
\end{lstlisting}
\caption{The trivial-channel prompt, verbatim. The model returns a chain of
terms as JSON; the harness matches each step against the law, bridges gaps of
at most three single applications, and assembles the Lean proof.}
\label{fig:prompt-trivial}
\end{figure*}

\begin{figure*}[t]
\begin{lstlisting}
You are constructing an infinite model for a magma law, to be checked by an
automated theorem prover.

The law is:  {law}
(the binary magma operation is written with the diamond; the law holds for all
inputs.)

This law has NO nontrivial FINITE model, so its model is necessarily infinite,
and it CANNOT be an arithmetic formula (any polynomial op over the integers
would descend to a finite model). Instead, propose a finite set of DEFINING
EQUATIONS for the operation — a presentation of the model — from which the law
follows and which is consistent with the carrier having two distinct elements.

Think of it as the completed rewrite system a Knuth-Bendix procedure would
converge to: a handful of equations that (a) entail the law, and (b) do not
force all elements equal.

Return ONLY a JSON object, nothing else:
  {"E": ["<equation>", "<equation>", ...]}
Each equation uses variables x,y,z,w and the diamond operator, in the form
LHS = RHS .
Example shape (NOT a solution):
  {"E": ["x ◇ (y ◇ x) = y", "(x ◇ x) ◇ x = x ◇ x"]}
Do NOT just restate the law, and do NOT write `x = y`.
\end{lstlisting}
\caption{The construction-channel prompt, verbatim. The model returns a finite
presentation $E$ as JSON, which the judge submits to the two checks of
Equation~(6).}
\label{fig:prompt-construct}
\end{figure*}

\section{Automated Baseline: Per-Configuration Results}
\label{app:perconfig}

The portfolio spans eight configurations: Vampire 5.0.1 in five modes, E 3.3.5
\citep{schulz2019faster} in two, and Twee 2.6.1. Table~\ref{tab:perconfig} reports how many distinct
laws each configuration resolves at each per-configuration budget, over the
4,141-law residual. The counts overlap, since several configurations resolve
the same law, so the column totals exceed the 114 laws resolved overall.

\begin{table*}[t]
\centering
\begin{tabular}{@{}lrrrrrrr@{}}
\toprule
Configuration & 30 & 60 & 120 & 300 & 600 & Total & Models \\
\midrule
twee/complete         & 69 & 5 & 4 & 3 & 5 & 86 & 4 \\
eprover/triv/auto     & 64 & 2 & 2 & 1 & 1 & 70 & 0 \\
vampire/triv/casc+lpo & 30 & 1 & 1 & 1 & 0 & 33 & 0 \\
vampire/triv/casc     & 24 & 1 & 0 & 1 & 0 & 26 & 0 \\
vampire/triv/discount & 22 & 0 & 0 & 0 & 0 & 22 & 0 \\
vampire/sat/otter+lpo & 13 & 1 & 0 & 0 & 0 & 14 & 0 \\
vampire/sat/otter+kbo & 12 & 1 & 1 & 0 & 0 & 14 & 0 \\
eprover/sat/satauto   &  0 & 0 & 0 & 0 & 0 &  0 & 0 \\
\bottomrule
\end{tabular}
\caption{Distinct laws resolved by each configuration at each per-configuration
budget in seconds, over the 4,141-law residual. Counts overlap across
configurations. ``Models'' counts laws resolved by exhibiting a nontrivial
model rather than by proving triviality.}
\label{tab:perconfig}
\end{table*}

Two features of the table bear on the main paper's claims. First, every
configuration concentrates its resolutions at the 30-second rung; no
configuration shows a profile in which added budget yields resolutions at a
steady rate. Second, all four models the submitted sweep recovers come from a
single configuration, \texttt{twee/complete}. The Vampire saturation modes
resolve 14 laws each, all by proving triviality. The construction side of the
portfolio therefore rested on one configuration.

The sweep exits a law's ladder as soon as any configuration resolves it, so the
number of laws reaching each rung declines: 4,141 at 30 seconds, then 4,050,
4,042, 4,037 and 4,033.

\section{LLM Failure Analysis}
\label{app:failures}

\begin{table}[t]
\centering
\begin{tabular}{@{}lrrrr@{}}
\toprule
LLM & Too strong & Too weak & Both fail & One check \\
\midrule
o3      & 23 & 2  & 0  & 25 \\
o4-mini & 11 & 4  & 10 & 15 \\
GPT-4.1 & 10 & 11 & 4  & 21 \\
\bottomrule
\end{tabular}
\caption{Final-round classification of all 25 hard-tier construction
submissions per LLM. ``Too strong'' entails the law but admits only
one-element models (passes check (a), fails (b)); ``too weak'' maintains two
distinct elements but fails to entail the law (passes (b), fails (a)). The last
column counts laws on which the LLM clears exactly one of the two checks.}
\label{tab:failures}
\end{table}

Each language model attempts all 25 hard-tier laws over three feedback rounds,
and no submission from any model passes both checks. Because a run ends either
in acceptance or after three rejections, every law leaves a final-round
submission, and Table~\ref{tab:failures} classifies all 75 of them by which of
the two checks that final submission satisfies.

The three classes partition the outcomes. A presentation is \emph{too strong}
when it entails the law but admits only one-element models: check~(a) succeeds
and check~(b) fails. It is \emph{too weak} when it keeps two elements distinct
but does not entail the law: check~(b) succeeds and check~(a) fails. A
submission failing both satisfies neither condition. The final column, counting
laws on which the model clears exactly one check, is therefore 25 minus the
both-fail count.

The trivial-side failures follow one dominant pattern at every setting, and it
is a different kind of failure. The submitted chains contain steps that are not
applications of the law: no substitution instance of either side rewrites one
term into the next, so the harness rejects the chain at the matching stage
rather than at the prover. The failure is in executing individual steps
soundly, not in planning a route between $a$ and $b$.

\section{Corpus Yield and Renewability}
\label{app:renew}

\begin{table}[t]
\centering
\begin{tabular}{@{}rrrr@{}}
\toprule
Order & Screened & Austin & Austin per 1{,}000 \\
\midrule
5 & 115   & 11  & 95.7$^{\ddagger}$ \\
6 & 1,418 & 29  & 20.5 \\
7 & 4,803 & 117 & 24.4 \\
8 & 4,138 & 105 & 25.4 \\
\bottomrule
\end{tabular}
\caption{Admissible-law yield by order, over all 10,474 screened candidates.
Density rises monotonically across the orders the extension engine generates,
so the supply does not thin as the corpus grows. $^{\ddagger}$Order 5 is
excluded from the trend: those laws come from ETP's curated enumeration rather
than a uniform draw from the generator.}
\label{tab:renewability}
\end{table}

Section~3.2 claims that the corpus has no terminal order and that fresh
instances can therefore be minted after any training cutoff. That claim is
worth checking against yield: a generator whose output thins as the order
grows would eventually stop supplying instances, whatever the recursion
permits in principle.

Table~\ref{tab:renewability} gives the yield of the screening pipeline broken
down by the order of the candidate law, over all 10,474 candidates screened.

Yield must be read as a density rather than a count. The raw number of Austin
laws dips from 117 at order 7 to 105 at order 8, but fewer order-8 candidates
were screened, so the dip measures how much was generated rather than how much
was there. Normalised to admissible laws per thousand screened candidates, the
yield rises across every order the extension engine produces: 20.5 at order 6,
24.4 at order 7, and 25.4 at order 8. Order 5 is excluded from the comparison
because those laws are ETP's curated enumeration rather than a uniform draw
from the generator, which is why its density is several times higher.

Counting only laws that open a new equivalence class gives 16.9, 19.4 and 16.2
per thousand across the same three orders. That density does not fall, so the
corpus continues to produce genuinely distinct problems rather than variations
on classes it has already covered; unlike the overall yield, it is not
monotonic, and we do not claim it is rising.

\section{Validation of the Automated Judge}
\label{app:judge}

Both channels of the judge run a control suite before any evaluation. The
suites are part of the released code and can be re-run with
\texttt{-{}-selftest}.

\paragraph{The trivial channel.} A submission is a Lean proof of a proposition
the judge generates, and two things must hold: the proof must elaborate at
exactly the generated type, and its axiom footprint must lie within the
allowlist $\{$\texttt{propext}, \texttt{Quot.sound},
\texttt{Classical.choice}$\}$. Anything outside that set is a failure, and
\texttt{sorryAx}, \texttt{Lean.ofReduceBool} and \texttt{Lean.trustCompiler}
are named explicitly as forbidden. The footprint is an allowlist rather than a
blocklist so that an unanticipated axiom fails closed.

Seven negative controls exercise the ways a submitter could avoid proving
anything, and each is rejected by a textual scan that runs before Lean is
invoked, since several of them would otherwise subvert Lean itself: a
\texttt{sorry}; a declared axiom asserting the goal; two forms of namespace
shadowing; redefining the law predicate as \texttt{True}; \texttt{native\_decide};
and naming the submitted theorem anything other than the identifier the judged
file elaborates. A positive control confirms the suite is not simply rejecting
everything: a reference answer passes the scan and compiles with an allowed
footprint.

\paragraph{The construction channel.} Three controls cover the three ways a
presentation can fail to encode a construction, matching the cases discussed in
Section~3.3. Submitting the law itself serves as the positive control: it
certifies exactly when bare saturation of $L \cup \{\,a \neq b\,\}$
terminates, which screening has already established for the seed law used
here. On the hard tier the same submission diverges at check~(b), so echoing
the law back does not produce a certificate there. The left projection $\{x \op y = x\}$ is too weak: it admits models
with distinct elements but does not entail the law, so check~(a) fails. The
collapse $\{x = y\}$ is too strong: it entails the law but admits only
one-element models, so check~(b) fails. The suite passes only when the first
certifies and the other two do not.

\section{Diversity of the Constructed Models}
\label{app:diversity}

Section~3.2 reports corpus size in equivalence classes, so that laws posing the
same problem are not counted twice. Logical equivalence is the natural
criterion, but it leaves a question open: two laws that are not equivalent may
still be satisfied by essentially the same construction, in which case the
class count would overstate how many distinct problems the corpus contains.

We test this directly, without a prover, by asking which laws each constructed
model satisfies. Every model built during screening is evaluated against every
one of the 262 laws proved Austin, giving 68,382 ordered pairs and no missing
models. A model transfers to a law when it satisfies that law as well as its
own.

The result is that construction classes and logical classes coincide exactly:
both number 195. The pairs on which transfer runs in both directions number
255, precisely the 255 pairs the prover proved equivalent, so mutual
model-satisfaction and logical equivalence pick out the same relation --- a
cross-check on the deduplication of Section~3.2 by a method that shares none of
its machinery. Transfer across class boundaries is rare and one-directional:
173 of 68,382 ordered pairs, or 0.25\%. A typical model satisfies only the law
it was built for, with a median coverage of one law, a mean of 3.67 and a
maximum of 19. No cell is undecided, and every model satisfies its own law.

The 195 classes are therefore construction-distinct and not merely logically
distinct. The models do not collapse onto a shared construction that a solver
could learn once and reapply.

\section{Computing Infrastructure and Cost}
\label{app:infra}

All prover experiments ran on a dual-socket AMD EPYC 7313 machine --- 2 sockets
$\times$ 16 cores with SMT disabled, giving 32 physical cores --- with 503\,GB
of RAM under Ubuntu 22.04.5 LTS. The portfolio sweep was sharded 32 ways, one
shard per physical core, so no configuration competed with a sibling
hyperthread for its time budget. Language-model experiments were run through
the OpenRouter API.

Table~\ref{tab:cost} reports token usage for every language-model run described
in the paper. The two certified-easy runs account for more than half of the
total, because each of the 63 laws may consume up to three feedback rounds and
because reasoning traces on that set are long: a solved law costs a median of
49,221 completion tokens. The construction runs are cheaper per law despite
resolving nothing, since a presentation is a short object to emit. GPT-4.1's
totals are an order of magnitude below the reasoning models' at comparable call
counts, which reflects the absence of a reasoning trace rather than a shorter
answer.

Files with more than 63 records contain repeated passes under identical
settings, appended in run order. The no-waypoint o3 file holds a first pass
over all 63 laws, on which four chains certify, followed by a second pass over
the 36 laws then unsolved, which adds two more; the six no-waypoint solutions
reported in the main text are the union of the two passes. The waypoint o3
file holds a first pass over all 63 laws, on which nine chains certify,
followed by two further passes over 32 of the unsolved laws, which add none.
The reproduction run is a separate file.

\begin{table}[t]
\centering
\begin{tabular}{@{}lrrr@{}}
\toprule
Run & Records & Completion & Prompt \\
\midrule
o3 cert-63, waypoints    & 127 & 3,094,531 & 164,044 \\
o3 cert-63, no waypoints & 99  & 3,289,573 & 130,963 \\
o3 low effort            & 9   & 152,309   & 20,724 \\
o3 reproduction          & 14  & 575,997   & 59,653 \\
o4-mini cert-63          & 63  & 974,972   & 147,300 \\
GPT-4.1 cert-63, waypoints    & 63 & 61,670 & 149,538 \\
GPT-4.1 cert-63, no waypoints & 63 & 71,736 & 139,620 \\
GPT-4.1 early run (95 laws)   & 100 & 192,466 & 269,443 \\
GPT-4.1 pilot (10 laws)  & 10  & 9,234     & 31,870 \\
o3 trivial, hard-25      & 25  & 1,226,978 & 53,475 \\
o3 construct, hard-25    & 25  & 1,120,628 & 144,122 \\
o4-mini construct        & 25  & 554,835   & 24,818 \\
GPT-4.1 construct        & 46  & 6,605     & 45,755 \\
\midrule
Total                    & 669 & 11,331,534 & 1,381,325 \\
\bottomrule
\end{tabular}
\caption{Token usage across all reported language-model runs. A solved
certified-easy law costs a median of 49,221 completion tokens.}
\label{tab:cost}
\end{table}

\end{document}